\pdfoutput=1

\documentclass[11pt]{article}

\usepackage[final]{acl}

\usepackage{times}
\usepackage{latexsym}

\usepackage[T1]{fontenc}

\usepackage[utf8]{inputenc}

\usepackage{microtype}

\usepackage{inconsolata}

\usepackage{graphicx}

\usepackage{hyperref}
\usepackage{subcaption}
\usepackage{booktabs}
\usepackage{amsmath,amssymb}
\usepackage{cleveref}
\usepackage{enumitem}
\usepackage{multirow}
\usepackage{xurl}
\usepackage[most]{tcolorbox}

\title{Language Model Re-rankers are Fooled by Lexical Similarities}

\def\authorsep{\hspace{0.3em}}

\author{Lovisa Hagström\textsuperscript{1,2} \authorsep Ercong Nie\textsuperscript{3,4} \authorsep Ruben Halifa\textsuperscript{5} \\ \textbf{Helmut Schmid\textsuperscript{3}} \authorsep \textbf{Richard Johansson\textsuperscript{1,2}} \authorsep \textbf{Alexander Junge\textsuperscript{5}} \medskip\\
\null\textsuperscript{1}Chalmers University of Technology \quad 
\null\textsuperscript{2}University of Gothenburg \\
\null\textsuperscript{3}LMU Munich \quad
\null\textsuperscript{4}Munich Center for Machine Learning \quad
\null\textsuperscript{5}amass technologies \\
\texttt{lovhag@chalmers.se}}

\begin{document}
\maketitle
\begin{abstract}
Language model (LM) re-rankers are used to refine retrieval results for retrieval-augmented generation (RAG). They are more expensive than lexical matching methods like BM25 but assumed to better process semantic information and the relations between the query and the retrieved answers. To understand whether LM re-rankers always live up to this assumption, we evaluate 6 different LM re-rankers on the NQ, LitQA2 and DRUID datasets. Our results show that LM re-rankers struggle to outperform a simple BM25 baseline on DRUID. Leveraging a novel separation metric based on BM25 scores, we explain and identify re-ranker errors stemming from lexical dissimilarities. We also investigate different methods to improve LM re-ranker performance and find these methods mainly useful for the more popular NQ dataset. Taken together, our work identifies and explains weaknesses of LM re-rankers and points to the need for more adversarial and realistic datasets for their evaluation.
\end{abstract}

\section{Introduction}

\begin{figure}[h]
    \centering
    \includegraphics[width=0.6\linewidth]{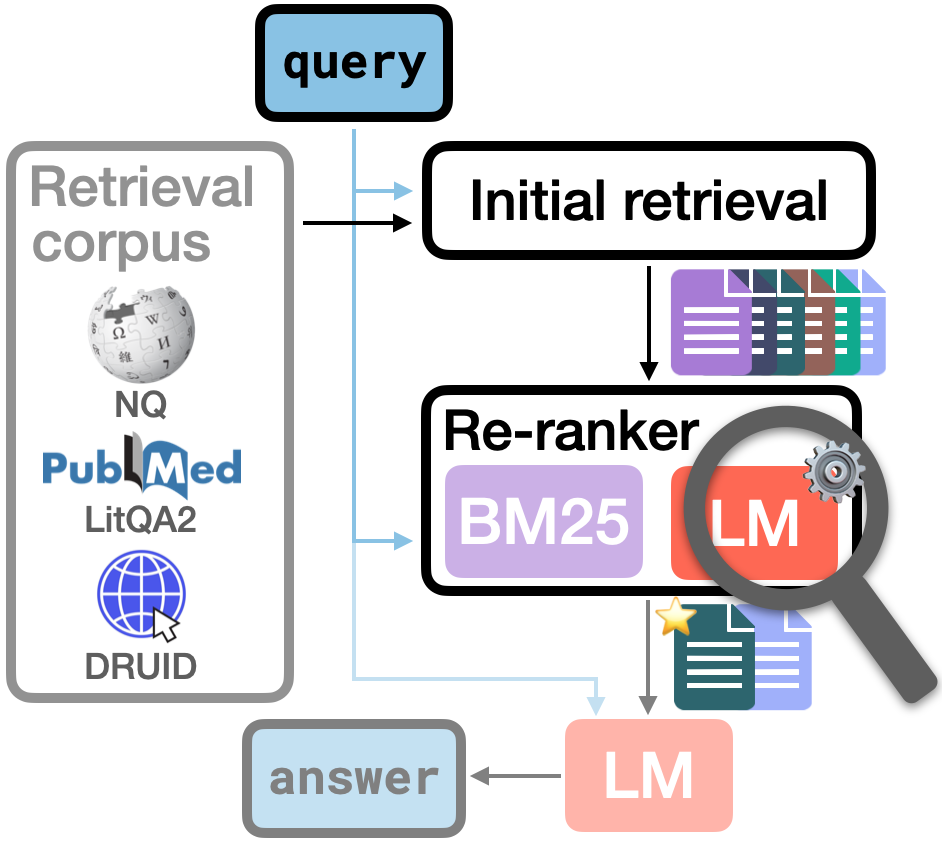}
    \caption{An overview of a RAG pipeline.}
    \label{fig:overview}
\end{figure}

Retrieval-augmented generation (RAG) is used to alleviate problems arising from imperfect parametric knowledge of language models (LMs) \citep{gao2023retrieval,vu-etal-2024-freshllms}. However, the efficiency of RAG hinges on the retrieval of useful information \citep{wang-etal-2024-searching}. To this end, LM re-rankers are increasingly used to provide more accurate retrieval results for RAG, superseding simpler methods based on keyword matching, such as BM25 (see \Cref{fig:overview}). While there are many benchmark results for LM re-rankers \citep{thakur2021beir,petroni-etal-2021-kilt}, little is known about when the computationally expensive LM re-rankers are worth the cost and whether they always can be expected to outperform simpler methods. 

In this paper, we evaluate LM re-rankers to better understand when they work well and when they fail to outperform less expensive alternatives.\footnote{Our code is available at \url{https://github.com/lovhag/rerankers-and-lexical-similarities}.} The contributions of this paper are as follows:
\begin{itemize}[leftmargin=*, itemsep=0pt]
\item We evaluate 6 LM re-rankers on the NQ, LitQA2 and DRUID datasets to compare re-ranker performance for scenarios of varying aspects of difficulty and domain. NQ is focused on generic QA, LitQA2 on scientific information extraction and DRUID on claim verification.
\item We explain variations in LM re-ranker performance using passage-query similarities, leveraging BM25 scores and our novel separation metric $D_S$. 
All LM re-rankers underperform on samples corresponding to low $D_S$ values and we tie these to high rates of \emph{distractors} (non-gold passages with high lexical similarity to the query) and \emph{lack of document context}.
\item We evaluate a set of methods for improving LM re-ranker performance, such as adding contextual information.
Our results show that while most methods work well on NQ, they are less effective for LitQA2 and DRUID.
\end{itemize}
Taken together, our paper identifies and measures novel aspects of difficulty for LM re-rankers; \emph{distractors} and \emph{lack of contextual information}.
These aspects are likely to occur in real-world scenarios relying on e.g. information retrieval from the web, such as in a fact-checking setting. Our work points to the need of more adversarial and real-world aligned evaluation datasets to better understand and address LM re-ranker deficiencies.

\section{Related Work}

The goal of using a re-ranker in an information retrieval context is to refine the outputs of an initial retrieval step based on a lexicographical or semantic database search. LM-based re-rankers are more expensive to run compared to simpler methods based on lexical matching, like BM25, but are expected to increase the performance of the overall retrieval system thanks to their semantic understanding~\citep{glass-etal-2022-re2g,li2023making}. 
\citet{sun-etal-2023-chatgpt} also showed how standard LLMs, like GPT-4, can be used as re-rankers.

Two popular benchmarks for re-rankers are the BEIR and KILT benchmarks by \citet{thakur2021beir,petroni-etal-2021-kilt}. 
Compared to our work, these benchmarks focus on high-level re-ranker performance and do not consider fine-grained aspects of difficulty for re-rankers. 

Similarly to our work, \citet{sturua2024jina} identify and investigate fine-grained aspects of difficulty for their \texttt{jina} models, of which one is \emph{misleading syntactic similarities}. This describes the case when passages with high syntactic similarity to the query are favoured over gold documents with lower syntactic overlap. Henceforth referred to as \emph{distractors}. 
\citet{wang-etal-2024-dapr} instead consider an aspect of difficulty related to \emph{missing document context}, for which a re-ranker may fail to identify a gold passage if its identification hinges on knowing that the passage comes from a relevant document or webpage. By prepending page titles to passages they were able to alleviate this issue on NQ. 

In contrast to these works, we expand on the analysis of distractors and missing document context to include multiple SOTA re-rankers, datasets from diverse domains and better tuned metrics. We also tie these aspects of difficulty to a more fundamental question of whether LM re-rankers are fooled by lexical similarities. To measure this, we develop a new metric which allows us to identify problematic samples.

\section{Method}

This section describes the re-rankers, datasets, metrics and alleviation methods investigated.

\subsection{Re-rankers}

We evaluate a wide cohort of LM re-rankers to enable comprehensive comparisons between different model types and sizes. Three closed-source LM re-rankers are evaluated: the industrial grade re-ranker Cohere\footnote{\texttt{rerank-english-v3.0} from \url{https://docs.cohere.com/v2/docs/models\#rerank}} (\texttt{Cohere}), a re-ranker based on GPT-4o (\texttt{GPT-4o}) and a lightweight re-ranker based on GPT-4o mini (\texttt{GPT-4o m}) \citep{sun-etal-2023-chatgpt} (\Cref{app:gpt}).\footnote{\texttt{gpt-4o-2024-08-06} and \texttt{gpt-4o-mini-2024-07-18}.}

We also evaluate three open-source re-rankers from Hugging Face: the large-scale LM re-ranker \texttt{bge-reranker-v2-gemma} (\texttt{BGE}), the lightweight re-ranker \texttt{jina-reranker-v1-turbo-en} (\texttt{Jina turbo}) and \texttt{jina-reranker-v2-base-multilingual} (\texttt{Jina base}), a larger re-ranker from the same model family. As a baseline, we consider BM25 scores, leveraging lexical matching, similar to TF-IDF \citep{bm25s}. See \Cref{app:runtimes} to get a rough estimate of the runtime of each re-ranker.

\subsection{Evaluation datasets}

We evaluate the re-rankers on three datasets representative of different domains and aspects of difficulty: NQ, LitQA2 and DRUID. All datasets have already undergone an initial retrieval step and are thus suitable for the evaluation of re-rankers. Natural Questions (NQ) is a popular dataset for re-ranker evaluations with passages from Wikipedia pages \citep{kwiatkowski-etal-2019-natural}. LitQA2 measures the ability of a system to extract information from scientific literature \citep{litqa2}, containing a high rate of domain-specific biomedical language. LitQA2 can be expected to test the robustness to domain-shifts of LM re-rankers. DRUID (Dataset of Retrieved Unreliable, Insufficient and Difficult-to-understand contexts) contains fact-checked claims and corresponding potential evidence automatically retrieved from the web \citep{druid}. It can be expected to contain more noisy passages and to test the capability of re-rankers to identify relevant information for fact-checking. More details and examples can be found in \Cref{app:datasets}.

\subsection{Evaluation metrics}

We mainly use Precision@1 ($\mathrm{P@1}$) for our re-ranker evaluations to accommodate the small number of passages available in DRUID.\footnote{Metrics are defined by TREC in \url{https://trec.nist.gov/pubs/trec16/appendices/measures.pdf}.} 
To understand when LM re-rankers fail to outperform simpler methods, we also compare to alignment with BM25 relevance scores, as follows.
\begin{equation}
    \Delta \mathrm{P@1}(R) = \mathrm{P@1}(R) - \mathrm{P@1}_{\mathrm{BM25}}(R)
    \label{eq:delta-p}
\end{equation} 
Given re-ranker predictions $R$, $\mathrm{P@1}(R)$ denotes the score measured when document relevance is given by gold labels (default) and $\mathrm{P@1}_{\mathrm{BM25}}(R)$ when relevance is given by BM25 scores. Leveraging this metric, we can investigate whether re-rankers align with gold labels over BM25 scores, which corresponds to positive $\Delta \mathrm{P@1}$ values. Negative $\Delta \mathrm{P@1}$ values correspond to when the re-ranker predictions align more with BM25 scores over gold labels. 

\subsection{Gold from similar separation metric}

To better understand why and when re-rankers fail to identify gold passages in a document, we define a \emph{gold-from-similar separation metric} $D_S$ for a given text similarity measure $S$. Given a query $q$, a set of passages $\mathbf{p}=\{p_1, ..., p_n\}$ and corresponding gold labels $\mathbf{y}$ indicating whether a passage $p_i$ is gold ($y_i=1$) or not ($y_i=0$), we compute the metric $D_S$ by subtracting the maximal similarity of the non-gold standard passages from the maximal similarity of the gold standard passages:
\begin{equation}
    D_S (q, \mathbf{p}, \mathbf{y}) = \max_{i:\ y_i=1} S(q, p_i) - \max_{i:\ y_i=0} S(q, p_i)
    \label{eq:metric}
\end{equation}
This metric indicates whether the most similar gold standard passage
    is more or less similar to the query than the most similar non-gold standard passage.

We assume there to exist at least one gold passage per ($q$, $\mathbf{p}$) sample.
The similarity measure $S$ can be any measure of choice that takes two documents as input. A larger value of $S$ should signify greater similarity between the two documents.

\subsection{Alleviation methods}\label{sec:alleviation}

We investigate two known methods previously shown to improve re-ranker performance: prepending page titles (\texttt{Prepend titles}) \citep{wang-etal-2024-dapr} and incorporating contextual information generated by GPT-4o mini (\texttt{Incorporate context}).\footnote{\url{https://www.anthropic.com/news/contextual-retrieval}} 
Prepending titles is quite straightforward for NQ and LitQA2, while the more noisy webpage text in DRUID yields low-quality titles, with missing values and inaccuracies. The DRUID samples also lack complete contexts, barring the \texttt{Incorporate context} method.
We instead experiment with adjusting the re-ranker prompt to better suit the fact-checking setting represented by DRUID (\texttt{Prompt}) (\Cref{app:tuned-prompt}). 

\section{Results}

The zero-shot performance of the re-rankers considered in this paper are shown in \Cref{tab:all-precision-diff}. Additional results can be found in \Cref{app:all-recall-extra-res}. Based on these results, we reach the following conclusions.

\begin{table}[h]
\small
\centering
    \begin{tabular}{lrrr}
    \toprule
    \textbf{Re-ranker} & \textbf{NQ} & \textbf{LitQA2} & \textbf{DRUID} \\
    \midrule
    \multicolumn{2}{l}{\emph{Standard mode}} \\
    \midrule
    Cohere & $0.65\ {\scriptstyle (0.13)}$ & $0.76\ {\scriptstyle (0.08)}$ & $0.68\ {\scriptstyle (-0.21)}$ \\
    BGE & $0.68\ {\scriptstyle (0.17)}$ & $0.78\ {\scriptstyle (0.10)}$ & $0.73\ {\scriptstyle (-0.15)}$ \\
    Jina turbo & $0.56\ {\scriptstyle (0.08)}$ & $0.61\ {\scriptstyle (0.03)}$ & $0.69\ {\scriptstyle (-0.20)}$ \\
    Jina base & $0.68\ {\scriptstyle (0.15)}$ & $0.65\ {\scriptstyle (0.06)}$ & $0.65\ {\scriptstyle (-0.20)}$ \\
    GPT-4o m & $0.83\ {\scriptstyle (0.37)}$ & $0.51\ {\scriptstyle (0.10)}$ & $0.72\ {\scriptstyle (-0.10)}$ \\
    GPT-4o & $\mathbf{0.85}\ {\scriptstyle (0.40)}$ & $0.50\ {\scriptstyle (0.10)}$ & $0.73\ {\scriptstyle (-0.10)}$ \\
    BM25 & $0.46$ & $0.67$ & $0.66$ \\
    \midrule
    \multicolumn{2}{l}{\emph{Prepend titles}} \\
    \midrule
    Cohere & $0.77\ {\scriptstyle (0.23)}$ & $0.79\ {\scriptstyle (0.09)}$ & $0.71\ {\scriptstyle (-0.17)}$ \\
    BGE & $0.76\ {\scriptstyle (0.23)}$ & $\mathbf{0.80}\ {\scriptstyle (0.10)}$ & $0.74\ {\scriptstyle (-0.14)}$ \\
    Jina turbo & $0.69\ {\scriptstyle (0.16)}$ & $0.66\ {\scriptstyle (0.02)}$ & $0.71\ {\scriptstyle (-0.17)}$ \\
    Jina base & $0.78\ {\scriptstyle (0.24)}$ & $0.77\ {\scriptstyle (0.07)}$ & $0.69\ {\scriptstyle (-0.18)}$ \\
    GPT-4o m & $\mathbf{0.85}\ {\scriptstyle (0.34)}$ & $0.50\ {\scriptstyle (0.08)}$ & $0.72\ {\scriptstyle (-0.08)}$ \\
    GPT-4o & $\mathbf{0.85}\ {\scriptstyle (0.36)}$ & $0.51\ {\scriptstyle (0.07)}$ & $0.74\ {\scriptstyle (-0.06)}$ \\
    BM25 & $0.50$ & $0.70$ & $0.68$ \\
    \midrule
    \multicolumn{2}{l}{\emph{Incorporate context}} & & \multicolumn{1}{|l}{\emph{Prompt}} \\
    \midrule
    Cohere & $0.72\ {\scriptstyle (0.24)}$ & $0.69\ {\scriptstyle (0.06)}$ & $0.69\ {\scriptstyle (-0.18)}$
    \\
    BGE & $0.72\ {\scriptstyle (0.26)}$ & $0.70\ {\scriptstyle (0.05)}$ & $0.77\ {\scriptstyle (-0.06)}$
    \\
    Jina turbo & $0.62\ {\scriptstyle (0.15)}$ & $0.60\ {\scriptstyle (-0.04)}$ & $0.72\ {\scriptstyle (-0.17)}$ 
    \\
    Jina base & $0.74\ {\scriptstyle (0.27)}$ & $0.63\ {\scriptstyle (0.09)}$ & $0.72\ {\scriptstyle (-0.14)}$
    \\
    GPT-4o m & $0.80\ {\scriptstyle (0.37)}$ & $0.47\ {\scriptstyle (0.12)}$ & $0.79\ {\scriptstyle (-0.02)}$ \\
    GPT-4o & $0.81\ {\scriptstyle (0.38)}$ & $0.46\ {\scriptstyle (0.11)}$ & $\mathbf{0.83}\ {\scriptstyle (0.05)}$ \\
    BM25 & $0.44$ & $0.58$ & $0.68$ 
    \\
    \bottomrule
    \end{tabular}
\caption{$\mathrm{P@1}$ of all re-rankers. Values in (parenthesis) indicate $\Delta \mathrm{P@1}$ (\Cref{eq:delta-p}). Values in \textbf{bold} indicate top scores.}
\label{tab:all-precision-diff}
\end{table}
\paragraph{LitQA2 is generally easier and NQ generally more difficult.}
The majority of the LM re-rankers perform best on LitQA2, followed by DRUID and NQ. The only exceptions are the Jina models and GPT-4o models. The GPT-4o models likely struggle on LitQA2 due to token limitations. 

\paragraph{Large LM re-rankers struggle to outperform a BM25 baseline on DRUID.}
The best-performing re-ranker (BGE) outperforms the BM25 baseline by 10\% on DRUID. This is smaller than the 46\% on NQ (for GPT-4o) and 15\% on LitQA2 (for BGE).
We also note that the smaller Jina LM re-rankers clearly outperform the BM25 baseline on NQ while they perform worse than or equal to BM25 on LitQA2 and DRUID.

\paragraph{LM re-rankers align more with BM25 scores than gold labels on DRUID.} 
The $\Delta \mathrm{P@1}$ values are negative for all LM re-rankers on DRUID in \Cref{tab:all-precision-diff}, indicating that the re-rankers align more with BM25 scores than gold labels on DRUID. 

We note that while DRUID is \emph{easier} compared to NQ with respect to LM re-ranker accuracy, it is \emph{harder} with respect to how LM re-rankers struggle to outperform simpler methods like BM25. We hypothesise that DRUID provides a greater challenge in this sense as it contains passages from the web and popular claims that may have seen frequent discussion, increasing the rate of distractors.

\subsection{Query-passage similarities}

\begin{figure}[t]
    \centering
    \begin{subfigure}[t]{\linewidth}
        \centering
        \includegraphics[trim={0 1.15cm 0 0},clip,width=\linewidth]{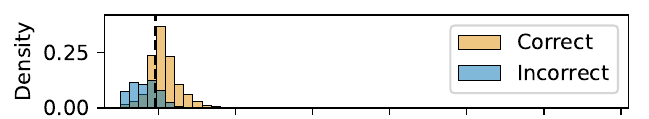}
        \caption{NQ}
    \end{subfigure}
    \begin{subfigure}[t]{\linewidth}
        \centering
        \includegraphics[trim={0 1.15cm 0 0},clip,width=\linewidth]{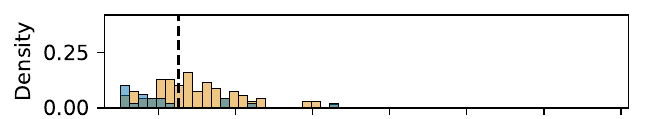}
        \caption{LitQA2}
    \end{subfigure}
    \begin{subfigure}[t]{\linewidth}
        \centering
        \includegraphics[width=\linewidth]{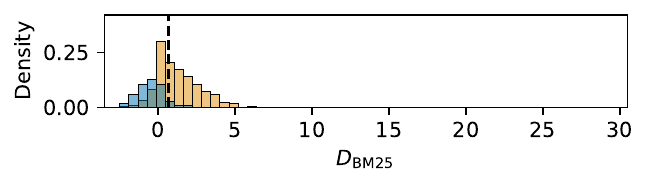}
        \caption{DRUID}
    \end{subfigure}
    \caption{Distribution of $D_{\mathrm{BM25}}$ (\Cref{eq:metric}) for NQ, LitQA2 and DRUID. Correctness is based on $\mathrm{P@1}$ of the BGE re-ranker. The dashed vertical lines indicate the mean values.}
    \label{fig:bm25_hist}
\end{figure}

To understand why LM re-rankers struggle to outperform BM25 on DRUID, we apply our separation metric $D_S$ to the passages in NQ, LitQA2 and DRUID and make comparisons to re-ranker precision. 
$D_{\mathrm{BM25}}$ results are found in \Cref{fig:bm25_hist} (results for other similarity metrics can be found in \Cref{app:all-recall-extra-res}). A summary of the distribution and corresponding re-ranker performance can be found in \Cref{tab:bm25_diffs}. To better understand the re-ranker performance on DRUID we also partition the dataset by $D_{\mathrm{BM25}}$ value and report the re-ranker scores in \Cref{tab:druid-partitioned-precision-diff}. Our conclusion is as follows.

\paragraph{LM re-rankers struggle to identify gold samples with markedly low BM25 scores.} The results in \Cref{fig:bm25_hist} show that LM re-rankers are generally good at identifying gold samples if they are sufficiently similar to the query. However, if the gold passage is too dissimilar to the query (corresponding to low $D_{\mathrm{BM25}}$ values), the LM re-rankers are prone to make mistakes.

We see how NQ and DRUID pose a greater challenge by including gold passages that are relatively dissimilar to the query. An inspection of some samples with low $D_{\mathrm{BM25}}$ scores in \Cref{app:d-samples} reveals a high rate of distractors and gold passages lacking document context. LitQA2 samples, on the other hand, have generally high $D_{\mathrm{BM25}}$ values and we hypothesise this makes the dataset easier for LM re-rankers. Seemingly, the domain-specific queries and passages of LitQA2 are less of a challenge compared to the lexical dissimilarities between gold passage and query in the other datasets.

\subsection{Alleviation methods}
The results from the investigations described in \Cref{sec:alleviation} can be found in \Cref{tab:all-precision-diff}. We reach the following conclusions.

\paragraph{Prepending page titles yields the greatest effects on NQ.}
Prepending page titles to the passages yields performance improvements for large LM re-rankers on NQ and unchanged performance on LitQA2 and DRUID. For LitQA2, this could be caused by the more distracting details from the scientific paper titles \citep{wang-etal-2024-dapr}. For DRUID it likely stems from the noisy webpage titles. Seemingly, the method of prepending page titles is more suitable for nicely formatted datasets, such as NQ. We also observe that the method of incorporating contexts is inferior to prepending page titles.

\paragraph{Adjusting the prompt yields significantly improved results for GPT-4o on DRUID.} \Cref{tab:all-precision-diff} shows how GPT-4o benefits the most from an adjusted prompt, indicating significance of prompt for the performance of LLMs as re-rankers.

\section{Conclusion}

Our paper identifies and explores an important weakness of LM re-rankers: they struggle to identify gold samples with markedly low BM25 scores. We hypothesise that real-world datasets like DRUID, with passages from the web, contain more distractors, resulting in gold samples with low BM25 scores. However, most current datasets for re-ranker evaluation fail to capture this aspect of difficulty and methods for improving LM re-ranker performance are less effective for the noisier LitQA2 and DRUID samples. Our work points to the need of more adversarial and real-world aligned datasets to better understand LM re-rankers and their weaknesses in realistic settings.

\section*{Limitations}

The datasets used in this study were not specifically designed to measure the preference of re-ranking models for similar over gold passages.
A dataset specifically curated for this purpose, potentially complemented by synthetically generated samples, would allow a deeper analysis of our research questions. We leave this for future work.

Our work only investigated a subset of the alleviation methods that exist for improving re-ranker performance. For example, there are also methods focused on adapting chunk sizes, and methods avoiding chunking all together. It would be interesting to also expand our analysis to incorporate additional alleviation methods.

\section*{Ethical Considerations}

There are no major ethical concerns related to our work on LM re-ranker performance. The datasets used and methods investigated are not associated with any ethical concerns.

\section*{Acknowledgments}
This work was supported by the Wallenberg AI, Autonomous Systems and Software Program (WASP) funded by the Knut and Alice Wallenberg Foundation. The computations were enabled by resources provided by the National Academic Infrastructure for Supercomputing in Sweden (NAISS) at Alvis partially funded by the Swedish Research Council through grant agreement no. 2022-06725. The work was also supported by compute credits from a Cohere For AI Research Grant.

\bibliography{anthology,custom}

\appendix

\section{Computational resources}

All open-source re-rankers are evaluated without fine-tuning on one T4, V100 or A100 Nvidia GPU per evaluation. The choice of GPU type depended on the model size (see \Cref{tab:runtimes} for detailed information on what GPU type was used for what model). The closed-source models were accessed via APIs so it is unclear as to exactly what GPU devices were involved. The total computational budget for the evaluations was about 50 GPU hours. 
 
\section{Use of AI assistants}

AI assistants like Copilot and ChatGPT were intermittently used to generate template code and rephrase sentences in the paper, etc. However, no complete paper sections or code scripts have been generated by an AI assistant. All generated text content has been inspected and verified by the authors. ChatGPT was also used and evaluated as a re-ranker in this work.

\section{Evaluation datasets}\label{app:datasets}

The evaluation datasets are described in further detail below. High-level statistics for the datasets can be found in \Cref{tab:datasets} and examples of samples from each dataset can be found in \Cref{tab:data-samples-nq,tab:data-samples-litqa,tab:data-samples-druid}. From each dataset we extract a set of \emph{questions}, corresponding \emph{passages} to choose between and corresponding \emph{gold labels} indicating whether a passage contains the answer to the given question or not.

\begin{table}[h]
    \small
    \centering
    \begin{tabular}{lrrrrr}
    \toprule
        Dataset & \#samples & \multicolumn{3}{c}{\#passages/sample} & \#gold  \\
        & & \scriptsize{mean} & \scriptsize{min} & \scriptsize{max} & /sample \\
    \midrule
        NQ & 3,759 & 16 & 4 & 244 & 1 \\
        LitQA2 & 124 & 145 & 33 & 359 & 1 \\
        DRUID & 875 & 4 & 2 & 5 & 2 \\
    \bottomrule
    \end{tabular}
    \caption{Statistics for the evaluation datasets. Exactly one gold passage is found per sample for NQ and LitQA2. DRUID samples may contain more than one gold passage.}
    \label{tab:datasets}
\end{table}

\begin{table*}[h]
    \small
    \centering
    \begin{tabular}{p{3cm} p{10cm} c}
    \toprule
        Question & Passages & Gold labels \\
    \midrule
        \multirow{2}{3cm}{when did hyderabad became a part of india?} & ``<H1> Hyderabad state (1948--56) </H1>'' & 0 \\
        & ``Jump to: navigation, search This article is about a State of the Indian Union 1948--1956 . For other uses, see Hyderabad (disambiguation).'' & 0 \\
        & ``<Table> <Tr> <Td colspan=3> Hyderabad State (1948 - 1956) </Td> </Tr> <Tr> <Td colspan=3> State of India </Td> </Tr> <Tr> <Td colspan=3> <Table> <Tr> <Td> \textbackslash u2190 </Td> <Td> 1948--1956 </Td> <Td> \textbackslash u2192 </Td> </Tr> </Table> </Td> </Tr> <Tr> <Td colspan=3> 1956 map of Southern India showing Hyderabad state in yellowish green . After the States reorganisation in 1956, regions west of the red and blue lines merged with Bombay and Mysore states respectively and the remaining part (Telangana) was merged with Andhra state to form Andhra Pradesh . </Td> </Tr> <Tr> <Td colspan=2> History </Td> <Td> </Td> </Tr> <Tr> <Td> </Td> <Td> Hyderabad State formed from Hyderabad Princely State </Td> <Td> 1948 </Td> </Tr> <Tr> <Td> </Td> <Td> Reorganized and renamed Andhra Pradesh </Td> <Td> 1956 </Td> </Tr> <Tr> <Td colspan=3> States of India since 1947 </Td> </Tr> </Table>'' & 0 \\
        & ``Hyderabad state until 1956'' & 0 \\
        & ``<P> Hyderabad State was a state in Independent India, formed after the accession of the princely state of Hyderabad into the Indian Union on 24 November 1949 . It existed from 1948 to 1956 . </P>'' & 1\\
    \bottomrule
    \end{tabular}
    \caption{Data sample from NQ.}
    \label{tab:data-samples-nq}
\end{table*}

\subsection{Natural Questions}
Natural Questions (NQ) by \citet{kwiatkowski-etal-2019-natural} is a popular dataset for re-ranker evaluations that contains real search engine queries and corresponding Wikipedia pages with the gold passage annotated. The gold passage annotators were instructed to identify the first paragraph on the Wikipedia page that contains the answer to the query, which means that there may be multiple unidentified gold passages for each query. To avoid issues stemming from this, we only retain all passages up to and including the gold passage as the retrieval corpora.

\paragraph{Chunking approach} The chunking is based on html elements, for which each passage is made out of one html element (e.g. a table <Table> or paragraph <P>), similarly to the approach used by the NQ authors to annotate gold passages. These passages are then matched to the annotated gold labels based on token indices.

\begin{table*}[h]
    \small
    \centering
    \begin{tabular}{p{3cm} p{10cm} c}
    \toprule
        Question & Passages & Gold labels \\
    \midrule
        \multirow{2}{3cm}{Neonatal male mice injected with NIF, a glycoprotein produced by a canine hookworm, show a significant reduction in microglial phagocytic capacity and engulfment of which neurotransmitter transporter? (A) VGlut2, (B) VGlut1, (C) VGlut3, (D) GAT1, (E) GAT2, (F) GAT3, or (G) not enough info?} & ``The incidences of neurodevelopmental disorders (NDDs) have been increasing in recent decades, suggesting a role for non-genetic environmental factors. Furthermore, sex is a significant risk factor for these disorders, with a strong male bias.'' & 0 \\
        & ``Air pollutant exposure during pregnancy or the first year of life is one of the most consistent environmental risk factors for NDDs. However, the associations of single environmental agents with NDDs have been relatively weak, and thus causality has been difficult to determine. Non-chemical stressors such as limited resources or social support of the mother can increase the vulnerability of the fetus to toxic exposures, which could explain why certain populations are disproportionately affected. In fact, neighborhood quality is a significant modifier of air pollution risk, suggesting that environmental and social stressors synergize to increase vulnerability to pollutant exposure, but how these exposures alter fetal brain development and affect offspring behavior is largely unknown.'' & 0 \\
        & ``Inflammatory events during pregnancy, such as maternal infection with bacteria or viruses, lead to maternal immune activation (MIA), which is linked to NDDs in offspring. Recent transcriptome-wide studies in postmortem brains of individuals diagnosed with an NDD have identified expression modules with enrichment of genes involved in neuroinflammatory function, with a particular dysregulation of microglial genes. Microglia are the primary immunocompetent cells of the brain and are exquisitely sensitive to perturbations of homeostasis and thus may be poised to act as immediate responders to environmental insults. Microglia are also essential regulators of activity-dependent synaptic remodeling during development, in which they prune inappropriate/weak synapses while sparing appropriate/strong connections. Importantly, transcriptome studies have found that immune changes co-occur with gene enrichment modules affecting synaptic function, suggesting the possibility that neuroimmune changes during development could lead to aberrant synapse development by altering microglial function.'' & 0 \\
        & ``A recent analysis found that MIA was more common in male children with ASD than female children, suggesting that a sex difference in response to maternal inflammation may be one mechanism that underlies increased male vulnerability. Furthermore, we and others have found sex differences in microglial development, maturation, and function, including an increased relative expression of microglial genes in male brains, compared with females. Interestingly, the microglial genes enriched in male brains are also implicated in ASD. Together these data point to a mechanism by which sexually dimorphic microglial responses to prenatal stressors could lead to aberrant brain development, primarily in males.'' & 0 \\
        & [...] & \\
        & ``In this experiment, WT neonatal male mice received bilateral microinjections of PBS or NIF (200 ng) into the ACC at P7, and brain tissue was collected 24 h later (Figure 7A). To confirm the effects of NIF on microglial phagocytic capacity, we quantified changes in the microglial lysosomal volume of CD68 (Figure 7B). As expected, microglia from animals microinjected with NIF had a significant reduction in the phagocytic index (~50\%) and a significant decrease in the total lysosomal content within each microglia (Figures 7C and 7D). To determine whether this reduction in CD68 impaired microglial interactions with VGlut2 synapses, we once again performed Imaris reconstructions and quantified the volume of VGlut2 within microglia (Figure 7E). Microglia from NIF-treated animals are significantly smaller (~25\%) than PBS control animals (Figure 7F); furthermore, this size reduction is accompanied by a significant decrease in the volume of internalized VGlut2 in microglia cells (Figure 7G). Last, we quantified the co-localization of VGlut2 and PSD95 and found that NIF-injected animals had about a 20\% increase in VGlut2+ synapses (Figure 7H). Thus, NIF injections at P7 effectively reduce microglial phagocytic capacity and engulfment of VGlut2, which induces an abnormal increase in VGlut2 synapse density.'' & 1 \\
        & [...] & \\
    \bottomrule
    \end{tabular}
    \caption{Data sample from LitQA2. ``[...]'' indicates that we are skipping across passages in the sample to save space.}
    \label{tab:data-samples-litqa}
\end{table*}

\subsection{LitQA2} 
LitQA2 by \citet{litqa2} measures the ability of a system to extract information from scientific literature. The dataset contains a high rate of domain-specific biomedical language compared to the more generic queries of NQ and can be expected to test the robustness to domain-shifts of LM re-rankers. The dataset consists of multiple-choice questions that are intended to be only answerable based on the full text, not on the abstract, of a given paper and nowhere else in the literature. 
PubMedCentral\footnote{\url{https://pmc.ncbi.nlm.nih.gov}} was used to scrape the full articles. Only 124 out of 200 samples were retained from this dataset as some articles were unavailable. We decided to include the dataset in the analysis despite the small sample size as this is the only high-quality dataset that enables evaluations of re-rankers for the biomedical domain. 

\paragraph{Chunking approach} The chunking is based on newlines, for which each passage is made out by a new paragraph. Passages can then be matched to the manually extracted gold passage via fuzzy matching, to get the gold labels for each passage.

\begin{table*}[h]
    \small
    \centering
    \begin{tabular}{p{2cm} p{4.5cm} p{4cm} c c}
    \toprule
        Question & Passages & Source & Gold labels & DRUID labels \\
    \midrule
        WikiLeaks has published the 1st list of black money holders in Swiss banks. & ``WikiLeaks has never published the list of Indians who have stashed their money in Swiss banks. Hence, the claim stands FALSE.'' & \url{https://factly.in/wikileaks-list-of-black-money-holders-in-swiss-bank-is-a-fake-one/} & 1 & refutes \\
        & ``Various posts on social media claim that WikiLeaks has released the "first list" of black money holders in Swiss Bank. The post is going viral on all social media platforms. DigitEye Team also received the message on its Whatsapp fact-checking number. The list contains 24 names \textbackslash u2014 Sonia Gandhi, A Raja, Rahul Gandhi, Sharad Pawar, P Chidambaram to name a few. All the money listed next to the names are figures in dollars. The first name on the alleged list is Congress leader Sonia Gandhi who it claimed to be holding more than \$56 billion. The numbers are not in chronological order and neither the names are in any set order. According to the alleged list, the lowest amount is held by P Chidambaram. [...] WikiLeaks has not published any report on the same on its website. The latest report was published in October 2019. WikiLeaks took to Twitter and tweeted about a similar list of Indian black money holders. In the 2011 tweet, it clarified that such list "never appeared on WikiLeaks".'' & \url{https://digiteye.in/viral-list-of-black-money-holding-accounts-in-swiss-bank-is-fake/} & 1 & refutes \\
        & ``INDIA/SWIZERLAND-- Black money trail: 2nd list of Indian Swiss accounts to be shared [...] TNN | Sep 14, 2011, 11.11AM IST http://timesofindia.indiatimes.com/ india/Black-money-trail-2nd-list-of-Indian-Swiss-accounts-to-be-shared/articleshow/9977871.cms NEW DELHI: A second list containing names of Indians, who have stashed black money in Swiss banks, will be shared by the Germans, Times Now reported.'' & \url{https://wikileaks.org/gifiles/docs/70/703306_india-swizerland-black-money-trail-2nd-list-of-indian-swiss.html} & 0 & insufficient \\
        & ``(See attached file: List of Black Money Holders from Wiki'' & \url{https://groups.google.com/g/yeida/c/V2gxTIXY-sQ} & 0 & insufficient \\
    \bottomrule
    \end{tabular}
    \caption{Data sample from DRUID. ``[...]'' inside a passage does not indicate additional information included to the re-ranker, it simply indicates that the passage was retrieved as snippets from a webpage, for which there is additional page content between the snippets.}
    \label{tab:data-samples-druid}
\end{table*}

\subsection{DRUID}
DRUID (Dataset of Retrieved Unreliable, Insufficient and Difficult-to-understand contexts) by \citet{druid} contains fact-checked claims and corresponding potential evidence pieces retrieved from the web. Each evidence piece has been annotated for whether it contains sufficient information to conclude whether the corresponding claim is true or false. The claims from the dataset are used as questions to the re-rankers and the collected DRUID passages corresponding to the given claim make out the passages for the query. Passages with sufficient information to reach a fact-check verdict, i.e. marked as `refuting' or `supporting', are considered gold and each sample corresponds to at least two potential passages from different webpages, of which at least one has to be gold and at least one not gold. The Cohere re-ranker was used for the automated retrieval of evidence pieces so the samples in DRUID can be expected to be more adversarial in the sense that they already have been pre-selected by a LM re-ranker (and then manually annotated for quality).

\paragraph{Chunking approach} The passages have already been chunked in a previous automated retrieval pipeline by the DRUID authors. Each passage is based on text snippets from a webpage, for which multiple snippets may have been extracted across the same webpage.

\clearpage

\section{Runtime comparison}\label{app:runtimes}

To exemplify the difference in efficiency between different re-rankers, we compare runtimes of the investigated re-rankers in \Cref{tab:runtimes}. Unfortunately, the models could not be run on the same devices due to space and other practical reasons.

\begin{table}[h]
    \small
    \centering
    \begin{tabular}{l r r}
    \toprule
    Re-ranker     & Runtime [mins] & Device \\
    \midrule
    Cohere & 15 & \emph{Cohere API} \\
    BGE & 42 & A100:1 \\
    Jina turbo & 3 & V100:1 \\
    Jina base & 80 & T4:1 \\
    GPT-4o m & 145 & \emph{OpenAI API} \\
    GPT-4o & 135 & \emph{Azure API} \\
    BM25     & 0.5 & MacBook Pro \\
    \bottomrule
    \end{tabular}
    \caption{Runtimes of the different re-rankers for getting scores corresponding to all samples from NQ (no prepended titles or context) on their corresponding devices. The MacBook Pro device is using a 2.3 GHz Quad-Core Intel Core i7.}
    \label{tab:runtimes}
\end{table}

\section{Implementation details of RankGPT}\label{app:gpt}
LLMs demonstrate strong capabilities in understanding long texts and handling complex tasks, making them suitable for use as re-rankers in passage re-ranking tasks. Building on the prompting strategies proposed by ~\citet{sun-etal-2023-chatgpt}, we explore the use of LLM-based re-rankers, specifically leveraging two advanced OpenAI models: GPT-4o (\texttt{gpt-4o-2024-08-06}) and GPT-4o mini (\texttt{gpt-4o-mini-2024-07-18}).
As illustrated in Figure~\ref{prompt_template}, the re-ranking process with LLMs is facilitated via prompting. Specifically, a set of text chunks, each assigned a unique identifier (e.g., \texttt{[1]},\texttt{[2]}) is provided as input to the LLM. 
The model is then instructed to reorder the chunks in descending order of relevance to a given query. 
The output is a ranked list of identifiers in a format such as \texttt{[3] > [4] > [1] > [2]}. 
Notably, this approach directly generates a ranking without calculating intermediate relevance scores.

For datasets such as \texttt{NQ} and \texttt{DRUID}, we apply this direct permutation generation strategy without modification. However, for the \texttt{LitQA2} dataset, the samples of which contain a significantly larger number of candidate chunks (an average of 145 per query), the token limitations of LLMs pose a challenge. 
To address this, we employ the sliding window strategy, following~\citet{sun-etal-2023-chatgpt}. 
This method processes the chunks iteratively, using a sliding window size $w$ and a step size $s$,  to re-rank the chunks in a back-to-first order. In our experiments on \texttt{LitQA2}, we set the window size to 20 and the step size to 2. However, we note that the GPT-4o re-ranker performance suffers on LitQA2 in spite of these adaptations.

\begin{figure}
    \centering

    \begin{tcolorbox}

\textbf{system:}

You are RankGPT, an intelligent assistant that can rank passages based on their relevancy to the query.\\

\textbf{user:}

I will provide you with \texttt{\{\{num\}\}} passages, each indicated by number identifier []. 
Rank them based on their relevance to query: \texttt{\{\{query\}\}}.\\

\textbf{assistant:}

Okay, please provide the passages.\\

\textbf{user:}

[1] \texttt{\{\{passage\_1\}\}}\\

\textbf{assistant:}

Received passage [1]\\

\textbf{user:}

[2] \texttt{\{\{passage\_2\}\}}\\

\textbf{assistant:}

Received passage [2]\\

(more passages) ...\\

\textbf{user}

Search Query: \texttt{\{\{query\}\}}. 

Rank the \texttt{\{\{num\}\}} passages above based on their relevance to the search query. The passages should be listed in descending order using identifiers, and the most relevant passages should be listed first, and the output format should be [] > [], e.g., [1] > [2]. Only response the ranking results, do not say any word or explain.
\end{tcolorbox}
    
    \caption{Prompt template for \texttt{GPT-4o} and \texttt{GPT-4o-mini} as re-rankers~\citep{sun-etal-2023-chatgpt}.}
    \label{prompt_template}
\end{figure}

\section{Adjusted prompt for DRUID}\label{app:tuned-prompt}

\begin{figure}
    \centering
    \includegraphics[width=1\linewidth]{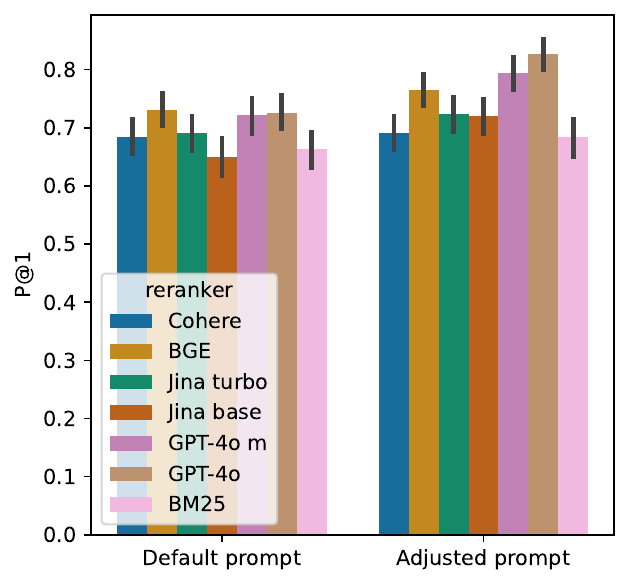}
    \caption{Re-ranker zero-shot alignment with gold labels on DRUID for different prompts.}
    \label{fig:druid-prompt-tuning}
\end{figure}

The prompts used for the prompt adjustment investigations for DRUID are as follows:
\begin{itemize}
    \item Default prompt: ``<claim>'' 
    \item Adjusted prompt: ``Is the following claim accurate?\textbackslash nClaimant: <claimant>\textbackslash nClaim: <claim>''
\end{itemize}
Here, ``<claim>'' and ``<claimant>'' are replaced by the corresponding values in DRUID. This prompt is tuned to adapt re-rankers to the fact-checking setting, as opposed to a QA setting. The results for these prompts can be found in \Cref{tab:all-precision-diff} and \Cref{fig:druid-prompt-tuning}.

\section{Additional re-ranker results}\label{app:all-recall-extra-res}
\begin{figure}[h]
    \centering
    \includegraphics[width=0.9\linewidth]{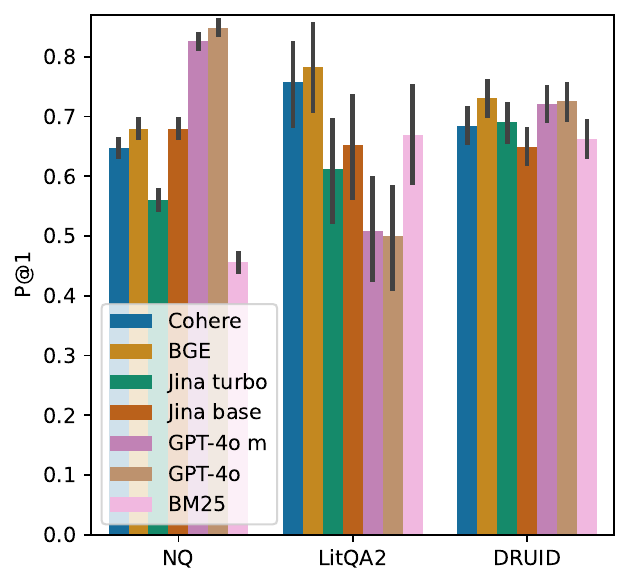}
    \caption{Re-ranker zero-shot alignment with gold labels for different datasets. The error bars indicate 95\% confidence intervals.}
    \label{fig:all-precision}
\end{figure}

\begin{figure}
    \centering
    \includegraphics[width=1\linewidth]{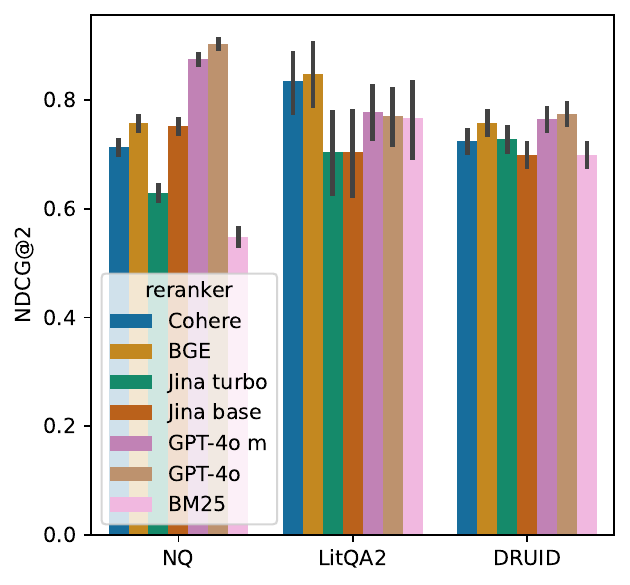}
    \caption{Re-ranker zero-shot alignment with gold labels for different datasets. The error bars indicate 95\% confidence intervals.}
    \label{fig:all-ndcg}
\end{figure}

Additional results corresponding to \Cref{tab:all-precision-diff} can be found in \Cref{fig:all-precision,fig:all-ndcg}. We also report additional $D_{\mathrm{BM25}}$ results in \Cref{tab:bm25_diffs,tab:druid-partitioned-precision-diff}.

\begin{table}[h]
    \small
    \centering
    \begin{tabular}{l l r r}
    \toprule
    \textbf{Dataset}     & \textbf{Partition} & \textbf{\% of data} & \textbf{P@1} \\
    \midrule
    NQ     & $D_{\mathrm{BM25}}<-0.5$ & 32 & 0.31 \\
           & $-0.5 \leq D_{\mathrm{BM25}}$ & 68 & 0.85 \\
    LitQA2     & $D_{\mathrm{BM25}}<-0.5$ & 31 & 0.47 \\
           & $-0.5 \leq D_{\mathrm{BM25}}$ & 69 & 0.92 \\
    DRUID     & $D_{\mathrm{BM25}}<-0.5$ & 20 & 0.24 \\
           & $-0.5 \leq D_{\mathrm{BM25}}$ & 80 & 0.85 \\
    \bottomrule
    \end{tabular}
    \caption{Re-ranker accuracy on the different datasets partitioned by $D_{\mathrm{BM25}}$ values. $\mathrm{P@1}$ is reported for \texttt{bge-reranker-v2-gemma}.}
    \label{tab:bm25_diffs}
\end{table}

\begin{table}[h]
    \small
    \centering
    \begin{tabular}{lrr}
    \toprule
    & \multicolumn{2}{c}{\textbf{DRUID}} \\
    \textbf{Re-ranker} & $D_{\mathrm{BM25}}<0.5$ & $0.5 \leq D_{\mathrm{BM25}}$ \\
    \midrule
    Cohere & $0.10\ {\scriptstyle (-0.78)}$ & $0.83\ {\scriptstyle (-0.07)}$ \\
    BGE & $0.24\ {\scriptstyle (-0.56)}$ & $0.85\ {\scriptstyle (-0.05)}$ \\
    Jina turbo & $0.13\ {\scriptstyle (-0.72)}$ & $0.83\ {\scriptstyle (-0.07)}$ \\
    Jina base & $0.18\ {\scriptstyle (-0.64)}$ & $0.77\ {\scriptstyle (-0.09)}$ \\
    GPT-4o m & $0.34\ {\scriptstyle (-0.41)}$ & $0.82\ {\scriptstyle (-0.02)}$ \\
    GPT-4o & $0.32\ {\scriptstyle (-0.40)}$ & $0.83\ {\scriptstyle (-0.02)}$ \\
    BM25 & $0.00$ & $0.83$ \\
    \bottomrule
    \end{tabular}
    \caption{Re-ranker zero-shot alignment with gold measured using $\mathrm{P@1}$ on DRUID partitioned by $D_{\mathrm{BM25}}$ values. Values in parenthesis indicate $\Delta \mathrm{P@1}$ (\Cref{eq:delta-p}).}
    \label{tab:druid-partitioned-precision-diff}
\end{table}

\begin{figure}
    \centering
    \begin{subfigure}[t]{\linewidth}
        \centering
        \includegraphics[trim={0 1.15cm 0 0},clip,width=\linewidth]{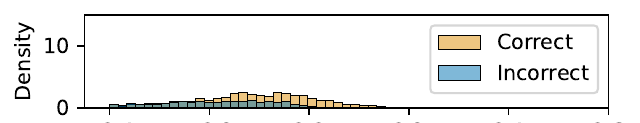}
        \caption{NQ}
    \end{subfigure}
    \begin{subfigure}[t]{\linewidth}
        \centering
        \includegraphics[trim={0 1.15cm 0 0},clip,width=\linewidth]{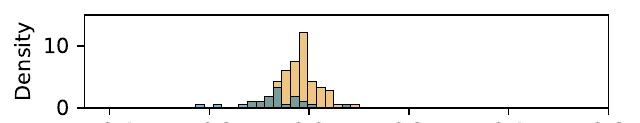}
        \caption{LitQA2}
    \end{subfigure}
    \begin{subfigure}[t]{\linewidth}
        \centering
        \includegraphics[width=\linewidth]{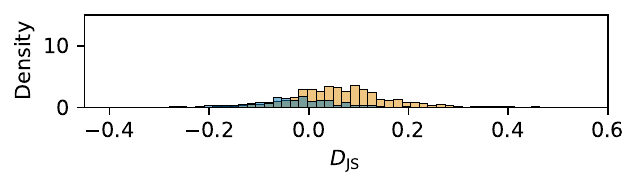}
        \caption{DRUID}
    \end{subfigure}
    \caption{$D_{\mathrm{JS}}$ (\Cref{eq:metric}) on NQ, LitQA2 and DRUID. Correctness is based on $\mathrm{P@1}$ for \texttt{bge-reranker-v2-gemma}.}
    \label{fig:js_hist}
\end{figure}

\begin{figure}
    \centering
    \begin{subfigure}[t]{\linewidth}
        \centering
        \includegraphics[trim={0 1.15cm 0 0},clip,width=\linewidth]{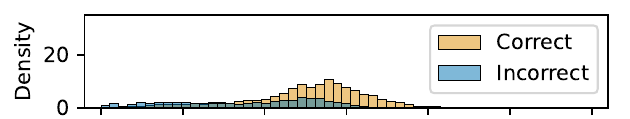}
        \caption{NQ}
    \end{subfigure}
    \begin{subfigure}[t]{\linewidth}
        \centering
        \includegraphics[trim={0 1.15cm 0 0},clip,width=\linewidth]{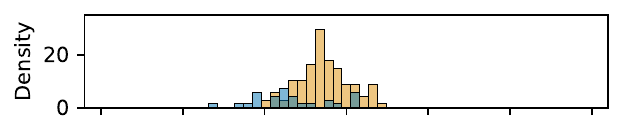}
        \caption{LitQA2}
    \end{subfigure}
    \begin{subfigure}[t]{\linewidth}
        \centering
        \includegraphics[width=\linewidth]{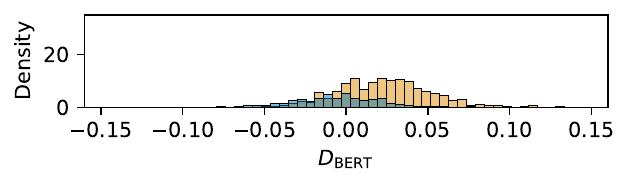}
        \caption{DRUID}
    \end{subfigure}
    \caption{$D_{\mathrm{BERT}}$ (\Cref{eq:metric}) on NQ, LitQA2 and DRUID. Correctness is based on $\mathrm{P@1}$ for \texttt{bge-reranker-v2-gemma}.}
    \label{fig:bert_hist}
\end{figure}

\begin{figure}
    \centering
    \begin{subfigure}[t]{\linewidth}
        \centering
        \includegraphics[trim={0 1.15cm 0 0},clip,width=\linewidth]{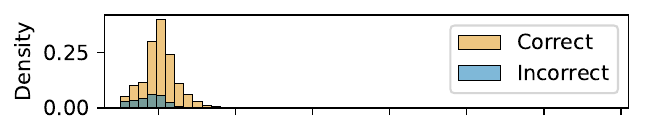}
        \caption{NQ}
    \end{subfigure}
    \begin{subfigure}[t]{\linewidth}
        \centering
        \includegraphics[trim={0 1.15cm 0 0},clip,width=\linewidth]{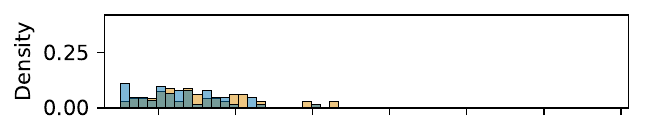}
        \caption{LitQA2}
    \end{subfigure}
    \begin{subfigure}[t]{\linewidth}
        \centering
        \includegraphics[width=\linewidth]{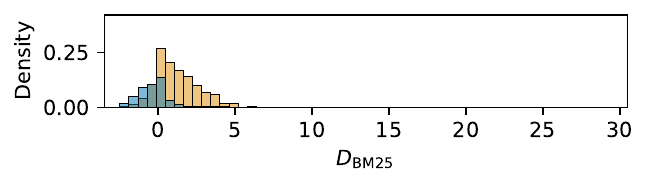}
        \caption{DRUID}
    \end{subfigure}
    \caption{$D_{\mathrm{BM25}}$ (\Cref{eq:metric}) on NQ, LitQA2 and DRUID. Correctness is based on $\mathrm{P@1}$ for GPT-4o.}
    \label{fig:bm25_hist_gpt4}
\end{figure}

\begin{figure}
    \centering
    \begin{subfigure}[t]{\linewidth}
        \centering
        \includegraphics[trim={0 1.15cm 0 0},clip,width=\linewidth]{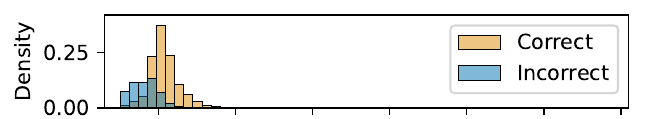}
        \caption{NQ}
    \end{subfigure}
    \begin{subfigure}[t]{\linewidth}
        \centering
        \includegraphics[trim={0 1.15cm 0 0},clip,width=\linewidth]{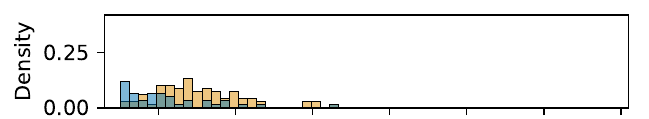}
        \caption{LitQA2}
    \end{subfigure}
    \begin{subfigure}[t]{\linewidth}
        \centering
        \includegraphics[width=\linewidth]{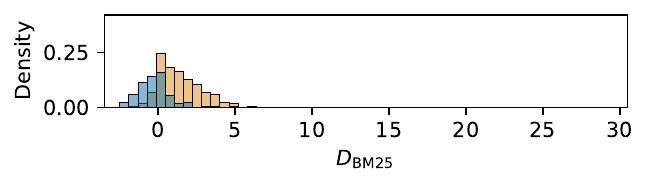}
        \caption{DRUID}
    \end{subfigure}
    \caption{$D_{\mathrm{BM25}}$ (\Cref{eq:metric}) on NQ, LitQA2 and DRUID. Correctness is based on $\mathrm{P@1}$ for \texttt{jina-reranker-v2-base-multilingual}.}
    \label{fig:bm25_hist_jina}
\end{figure}

Additional separation results for the similarity measures Jaccard similarity ($D_{\mathrm{JS}}$) and BERT score ($D_{\mathrm{BERT}}$) can be found in \Cref{fig:js_hist,fig:bert_hist}. $D_{\mathrm{BM25}}$ scores with correctness evaluated based on GPT-4o and Jina base scores can be found in \Cref{fig:bm25_hist_gpt4,fig:bm25_hist_jina}.

\section{Samples with different separation values}\label{app:d-samples}

\Cref{tab:subset-examples-nq,tab:subset-examples-litqa,tab:subset-examples-druid} contain samples from NQ, LitQA2 and DRUID with corresponding $D_{\mathrm{BM25}}$ values.

\begin{table*}[h]
    \scriptsize
    \centering
    \begin{tabular}{l p{2cm} p{5.5cm} p{5.5cm}}
    \toprule
    $D_{\mathrm{BM25}}$ & Question & Gold passage & Most similar passage \\
    \midrule
    -4.92 & who won the academy award for best original musical score? & \textcolor{Purple}{<Table> <Tr> <Th> Year </Th> <Th> Film </Th> <Th> Nominees </Th> </Tr> <Tr> <Td> (83rd) </Td> </Tr> <Tr> <Td> The Social Network </Td> <Td> Trent Reznor \& Atticus Ross </Td> </Tr> <Tr> <Td> How to Train Your Dragon </Td> <Td> John Powell </Td> </Tr> <Tr> <Td> Inception </Td> <Td> Hans Zimmer </Td> </Tr> <Tr> <Td> The King's Speech </Td> <Td> Alexandre Desplat </Td> </Tr> <Tr> <Td> 127 Hours </Td> <Td> A.R. Rahman </Td> </Tr> <Tr> <Td> 2011 (84th) </Td> </Tr> <Tr> <Td> The Artist </Td> <Td> Ludovic Bource </Td> </Tr> <Tr> <Td> The Adventures of Tintin </Td> <Td> John Williams </Td> </Tr> <Tr> <Td> Hugo </Td> <Td> Howard Shore </Td> </Tr> <Tr> <Td> Tinker Tailor Soldier Spy </Td> <Td> Alberto Iglesias </Td> </Tr> <Tr> <Td> War Horse </Td> <Td> John Williams </Td> </Tr> <Tr> <Td> 2012 (85th) </Td> </Tr> <Tr> <Td> Life of Pi </Td> <Td> Mychael Danna </Td> </Tr> <Tr> <Td> Anna Karenina </Td> <Td> Dario Marianelli </Td> </Tr> <Tr> <Td> Argo </Td> <Td> Alexandre Desplat </Td> </Tr> ... </Td> </Tr> </Table>} & <P> The Academy began awarding movies for their scores in 1935 . The category was originally called Best Scoring . At the time, winners and nominees were a mix of original scores and adaptations of pre-existing material . Following the controversial win of Charles Previn for One Hundred Men and a Girl in 1938, a film without a credited composer that featured pre-existing classical music, the Academy added a Best Original Score category in 1939 . In 1942, the distinction between the two Scoring categories changed slightly as they were renamed to Best Music Score of a Dramatic Picture and Best Scoring of a Musical Picture . This marked the first time the category was split into separate genres, a distinction that technically still lasts today, although there haven't been enough submissions for the musical category to be activated since 1985 . From 1942 to 1985, musical scores had their own category, with the exception of 1958, 1981 and 1982 . During that time, both categories had many name changes: </P> \\
    \midrule
    -3.68 & tumhi ho bandhu sakha tumhi cast real name? & \textcolor{Purple}{<Ul> <Li> Chandni Bhagwanani as Sanjana Ajay Pethewala </Li> <Li> Sreejita De as Shreya Bhushan Pethewala </Li> <Li> Kabeer K as Ajay Pethewala </Li> <Li> Neil Bhatt as Bhushan Trilokchand Pethewala </Li> <Li> Dimple Jhangiani as Avni Pethawala </Li> <Li> Lavina Tandon as Shaina </Li> ... </Ul>} & <P> The show began with the working title Pethawala before being named Tum Hi Ho Bandhu Sakha Tumhi . The show ended due to low trp ratings . </P> \\
    \midrule
    -0.33 & when did the movie karate kid come out? & <P> Jaden Christopher Syre Smith (born July 8, 1998) is an American actor, rapper, singer and songwriter . He is the son of Jada Pinkett Smith and Will Smith . Jaden Smith's first movie role was with his father in the 2006 film The Pursuit of Happyness . He also acted in the 2008 remake of The Day the Earth Stood Still and the 2010 remake of The Karate Kid, and was in the 2013 film After Earth with his father . </P> & [same as gold] \\
    \midrule
    5.84 & who said i think there is a world market for maybe five computers? & <P> Although Watson is well known for his alleged 1943 statement, "I think there is a world market for maybe five computers," there is scant evidence he said it . Author Kevin Maney tried to find the origin of the quote, but has been unable to locate any speeches or documents of Watson's that contain this, nor are the words present in any contemporary articles about IBM . One of the very first attributions may be found in The Experts Speak, a book written by Christopher Cerf and Victor S. Navasky in 1984, however Cerf and Navasky just quote from a book written by Morgan and Langford, Facts and Fallacies . Another early article source (May 15, 1985) is a column by Neil Morgan, a San Diego Evening Tribune writer who wrote: "Forrest Shumway, chairman of The Signal Cos., doesn't make predictions . His role model is Tom Watson, then IBM chairman, who said in 1958:' I think there is a world market for about five computers ."' The earliest known citation on the Internet is from 1986 on Usenet in the signature of a poster from Convex Computer Corporation as "' I think there is a world market for about five computers'--Remark attributed to Thomas J. Watson (Chairman of the Board of International Business Machines), 1943". All these early quotes are questioned by Eric Weiss, an editor of the Annals of the History of Computing in ACS letters in 1985 . </P> & [same as gold] \\
    \bottomrule
    \end{tabular}
    \caption{Examples of samples from NQ with relatively high and low $D_{\mathrm{BM25}}$ values. Passages lacking document context are marked in \textcolor{Purple}{purple}. Passages containing distractors are marked in \textcolor{PineGreen}{green} with the distracting terms in \textcolor{PineGreen}{\textbf{bold}}.}
    \label{tab:subset-examples-nq}
\end{table*}

\begin{table*}[h]
    \scriptsize
    \centering
    \begin{tabular}{l p{2.7cm} p{5.15cm} p{5.15cm}}
    \toprule
    $D_{\mathrm{BM25}}$ & Question & Gold passage & Most similar passage \\
    \midrule
    -5.97 & How long do mouse neurons survive following CRISPR inactivation of HSPA5? (A) 14 days, (B) 3 days, (C) 5 days, (D) 10 days, (E) 28 days, or (F) not enough info? & We selected Hspa5, a top hit that was not previously identified as a hit in iPSC-derived neurons, for individual validation. In mouse embryonic fibroblasts expressing CRISPRi machinery, we confirmed that an sgRNA targeting Hspa5 (sgHspa5) suppresses expression of the endogenous Hspa5 transcript (Fig. 5a). In primary neurons cultured from conditional CRISPRi mice, AAVs delivering sgHspa5 led to marked Cre-dependent neuronal death within 2 weeks of expression (Fig. 5b,c). Furthermore, injection of this sgRNA into neonatal mice led to a severe motor phenotype after approximately 2 weeks in mice co-expressing hSyn1-Cre, but not the sgRNA alone (Supplementary Videos 1 and 2), and the brains from mice with sgHspa5 + hSyn1-Cre were markedly smaller in size relative to sgHspa5-only littermates (Fig. 5d). This confirms the capability of our platform to uncover neuron-essential genes. & \textcolor{PineGreen}{For \textbf{mouse} primary \textbf{neurons} transduced with AAV, live imaging was performed every other day using an ImageXpress Micro Confocal HT.ai High-Content Imaging System (Molecular Devices). The imaging chamber was warmed to 37°C and equilibrated with 5\% CO2. The system used an Andor Zyla 4.5 camera with a Plan Apo ×10/0.45NA objective lens, an 89 North LDI laser illumination unit, 10-500 ms exposure time, 1×1 binning, and 10\% laser intensity using 405-nm, 475-nm, and 555-nm lasers, running MetaXpress (version 6.7.1.157). Resulting images were imported into Cell Profiler (version 4.2.1)28 and analyzed using a custom pipeline. hSyn1-Cre-GFP+ nuclei were segmented using the ‘IdentifyPrimaryObjects’ module, with expected diameter 8-40 pixels, using an Adaptive threshold (size 50) and the Minimum Cross-Entropy method, with a 1.5 smoothing scale, 1.0 correction factor, and lower- and upper-bound threshold at 0.435 and 1, respectively. Segmented objects were exported, and counted in each field, then summed across all fields within a well to calculate the number of objects per well (n=29 fields per well, n=4 wells per condition), using a custom R script. This was repeated for each timepoint. Data was normalized to fluorescent intensity at \textbf{day} 8 (as before that day, fluorescence intensity increased linearly with time in all channels as cells manufactured fluorescent proteins) and percentage change was calculated for each well from \textbf{day} 8, for subsequent timepoints through \textbf{day} 16.} \\
    \midrule
    11.62 & Based on whole genome bisulfite sequencing data (WGBS) from publicly available datasets (the ROADMAP epigenome project and the ENCODE data portal), what is the relationship between DNA methylation patterns between introns and exons (after excluding consideration of the first intron and first exon)? (A) There are no significant differences, (B) Introns have more DNA methylation, (C) Exons have more DNA methylation, (D) Neither introns nor exons can be methylated, (E) only areas very close to the transcription start site, or (F) not enough info? & Further, we considered a possible association between these gradients in DNA methylation, and the mutation risk in comparing exonic versus intronic DNA, in light of reports of subtly different mutation rates and subtly different DNA methylation in exons versus introns . We checked the DNA methylation level of exons and introns, separately for each exon/intron in sequence, for a representative gene set (middle tertile of genes by length, and middle tertile in expression level). While methylation in the first exon was substantially lower compared to the first intron, consistent with the exon's more 5' positioning, the DNA methylation levels across the subsequent introns and exons were highly similar (Supplementary Figure S7). Thus, in human WGBS data, after accounting for 5' gene end hypomethylation, we see no notably different DNA methylation in the exonic versus intronic loci, and if there are any differences between introns and exons in mutation rates, these do not stem from different DNA methylation. & [same as gold] \\
    \bottomrule
    \end{tabular}
    \caption{Examples of samples from LitQA2 with relatively high and low $D_{\mathrm{BM25}}$ values. Passages lacking document context are marked in \textcolor{Purple}{purple}. Passages containing distractors are marked in \textcolor{PineGreen}{green} with the distracting terms in \textcolor{PineGreen}{\textbf{bold}}.}
    \label{tab:subset-examples-litqa}
\end{table*}

\begin{table*}[h]
    \scriptsize
    \centering
    \begin{tabular}{l p{2cm} p{5.5cm} p{5.5cm}}
    \toprule
    $D_{\mathrm{BM25}}$ & Question & Gold passage & Most similar passage \\
    \midrule
    -4.71 & "Before the pandemic, just over 40,000 were on continuing UI claims. Now, there are well over 100,000 on state or federal UI benefits." & Department of Workforce Development datashows that in the week ending on March 7, 2020, there were 41,015 unemployment claims statewide. For the week of May 22, 202, there were 127,745 claims. & \textcolor{PineGreen}{Lisa Subeck stated on February 16, 2024 in X, formerly Twitter: "The United States is an outlier, one of only about half a dozen countries, without any guarantee of paid leave for new parents and/or other health care needs." Tim Kaine stated on March 15, 2022 in a tweet.: "Virginia women are paid 80 cents for every dollar paid to Virginia men." Mandela Barnes stated on May 23, 2021 in Twitter: "It’s been over 50 years since minimum (wage) and inflation parted ways, then over a decade since the federal minimum went up at all." Glenn Grothman stated on June 8, 2021 in Twitter: "We have a record 9.3 million job openings in the U.S." Mark Born stated on June 2, 2021 in Twitter: \textbf{"Before the pandemic, just over 40,000 were on continuing UI claims.} Mandela Barnes stated on May 23, 2021 in Twitter: "Since 1978, CEO compensation rose over 1,000\% and only 11.9\% for average workers." Joe Biden stated on April 15, 2020 in comments at a virtual town hall meeting: "Until this week, they [OSHA] weren’t even enforcing these guidelines [for coronavirus]. [...] Mark Born stated on June 2, 2021 in Twitter: \textbf{"Before the pandemic, just over 40,000 were on continuing UI claims.}} \\
    \midrule
    -3.59 & Claims that President George Washington once said, "Government is not reason; it is not eloquence; it is force. Like fire, it is a dangerous servant and a fearful master." & \textcolor{Purple}{There is no record of Washington ever making this statement.} & \textcolor{PineGreen}{FACT CHECK: Did George Washington Call Government \textbf{‘A Dangerous Servant And A Fearful Master’}? An image shared on Facebook claims that President George Washington once said, \textbf{"Government is not reason; it is not eloquence; it is force.} [...] According to the website Quote Investigator, the depiction of government as \textbf{"a dangerous servant and a fearful master"} is reminiscent of a centuries-old saying about water and fire. "Water is a very good seruaunt, but it is a cruell mayster," reads an excerpt from 1562.} \\
    \midrule
    3.88 & The Police Service of Northern Ireland (PSNI) are to pilot a Snapchat social media platform initiative to monitor social mitigation compliance in Northern Ireland. & The PSNI have no plans to introduce any monitoring scheme on any social media platform. Complaints about social mitigation compliance can be registered on the PSNI website. A claim was published on social media, that the Police Service of Northern Ireland (PSNI) is "to roll out a new ‘Snap-fish’ pilot scheme" on the Snapchat social media platform "to help catch individuals not adhering to social distancing, social bubbles and gathering more than six people" (often referred to as "social mitigation compliance"). [...] "The Police Service of Northern Ireland has no plans to introduce a ‘snap-fish’ scheme … nor indeed any new social media platforms around the enforcement of COVID-19 restrictions." & [same as gold] \\
    \midrule
    5.88 & Families of the deceased persons to be given an assistance of 4 lakh rupees, up from 2.5 lakh rupees. & Is assistance of Rs. 4 lakh being provided for families of deceased persons? The third claim is that ‘families of the deceased persons to be given an assistance of 4 lakh rupees, up from 2.5 lakh rupees’. It is true that the revised norms of assistance from the SDRF increase the assistance per deceased persons to Rs. 4 lakh from the existing Rs. 1.5 lakh per person. It has to be noted that is not for farmers alone, but for any deceased person during a notified natural disaster. & [same as gold] \\
    \bottomrule
    \end{tabular}
    \caption{Examples of samples from DRUID with relatively high and low $D_{\mathrm{BM25}}$ values. Passages lacking document context are marked in \textcolor{Purple}{purple}. Passages containing distractors are marked in \textcolor{PineGreen}{green} with the distracting terms marked in \textcolor{PineGreen}{\textbf{bold}}.}
    \label{tab:subset-examples-druid}
\end{table*}

\end{document}